\documentclass[10pt,twocolumn,letterpaper]{article}

\usepackage{cvpr}      

\usepackage{makecell}

\definecolor{cvprblue}{rgb}{0.21,0.49,0.74}
\usepackage[pagebackref,breaklinks,colorlinks,allcolors=cvprblue]{hyperref}

\def\paperID{5456} 
\def\confName{CVPR}
\def\confYear{2025}

\title{StageDesigner: Artistic Stage Generation for Scenography via Theater Scripts}

\author{Zhaoxing Gan$^1$, ~
Mengtian Li\footnotemark[2] $^,$$^1$$^,$$^2$, ~
Ruhua Chen$^3$, ~
Zhongxia Ji$^3$, \\
Sichen Guo$^3$, ~
Huanling Hu$^1$, ~
Guangnan Ye\footnotemark[2] $^,$$^1$, ~
Zuo Hu$^3$, ~
\\
$^1$Fudan University, ~
$^2$Shanghai University, ~
$^3$Shanghai Theatre Academy \\
{\tt\small zxgan23@m.fudan.edu.cn, mtli@shu.edu.cn, yegn@fudan.edu.cn}\\
{\tt\small \{chenruhua, zhongxia.ji, huzuo\}@sta.edu.cn}
}

\begin{document}
\twocolumn[{%
\maketitle
\begin{center}
    \centering
    \captionsetup{type=figure}
    \includegraphics[width=1\textwidth]{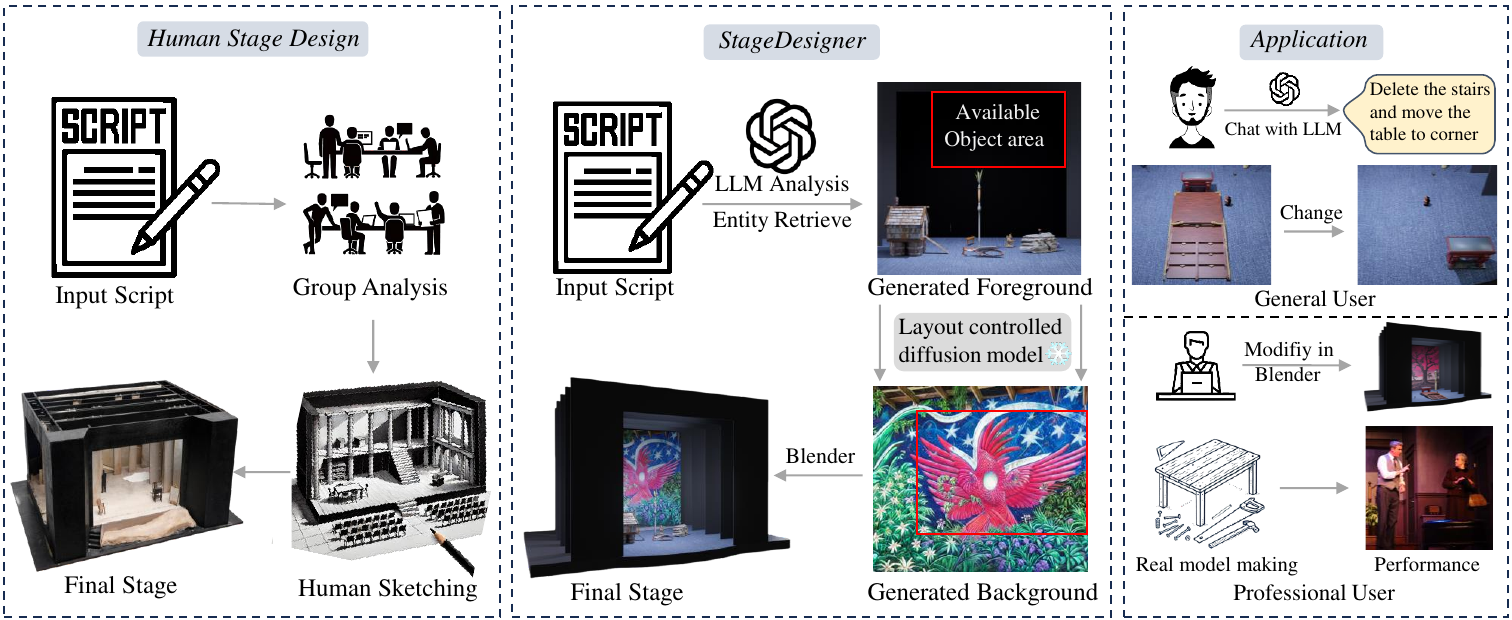}
    \captionof{figure}{Human Stage Design: Traditional stage design requires manual script analysis, collaborative planning, and model creation, making it time-consuming and expertise-dependent. StageDesigner: StageDesigner automates stage design by transforming scripts into 3D foreground elements and then placing background elements in unobstructed areas within the audience’s line of sight. 
    Application: General users can refine stage designs through conversational edits by feeding results back into the LLM, while professionals can import designs into Blender for detailed adjustments, real model creation, and performance preparation.}
    \label{fig: demo}
\end{center}%
}]
\begin{abstract}
In this work, we introduce \textbf{StageDesigner}, the first comprehensive framework for artistic stage generation using large language models combined with layout-controlled diffusion models. Given the professional requirements of stage scenography, StageDesigner simulates the workflows of seasoned artists to generate immersive 3D stage scenes. Specifically, our approach is divided into three primary modules: \textit{Script Analysis}, which extracts thematic and spatial cues from input scripts; \textit{Foreground Generation}, which constructs and arranges essential 3D objects; and \textit{Background Generation}, which produces a harmonious background aligned with the narrative atmosphere and maintains spatial coherence by managing occlusions between foreground and background elements.
Furthermore, we introduce the \textbf{StagePro-V1} dataset, a dedicated dataset with 276 unique stage scenes spanning different historical styles and annotated with scripts, images, and detailed 3D layouts, specifically tailored for this task. Finally, evaluations using both standard and newly proposed metrics, along with extensive user studies, demonstrate the effectiveness of StageDesigner. Project can be found at: \normalsize \url{https://deadsmither5.github.io/2025/01/03/StageDesigner/}
\end{abstract}
    
\section{Introduction}
Artistic stage design for theater and performance art is a complex task, transforming textual narratives into immersive visual environments~\cite{essin2012stage,wolf2014scene,stern2021stage}. Traditional scenography requires significant time and expertise to accurately capture a theater script’s intended mood, spatial relationships, and thematic depth.
Therefore, developing an AI-driven, user-friendly approach to artistic stage design could enable artists to create high-quality stages more efficiently. However, despite recent progress in 3D indoor scene synthesis~\cite{wang2021sceneformer,paschalidou2021atiss,feng2024layoutgpt,fu2025anyhome,yang2024holodeck} and text-to-image generation~\cite{rombach2022high,zhang2023adding}, little attention has been given to the unique challenges of artistic stage generation, where spatial coherence, thematic alignment, and narrative fidelity are crucial.

To address those challenges, we present \textbf{StageDesigner}, the first AI framework specifically designed for automated artistic stage generation based on theater scripts. StageDesigner employs a novel pipeline that leverages large language models (LLMs) and layout-controlled diffusion models to generate 3D foreground elements and atmospherically aligned backgrounds from script. The framework interprets script elements, arranges foreground entities, and generates cohesive background images, effectively bridging the gap between textual descriptions and visual scenography. StageDesigner stands out from existing scene generation methods by addressing the spatial and thematic intricacies unique to artistic stage settings, such as audience perspective considerations and positioning entities to avoid obstructing key visual elements.

Specifically, StageDesigner contains two key components, including the Script Analysis Module and the Foreground Projection Module. Script Analysis Module extracts relevant entities, spatial relationships, and thematic elements from a script. This module ensures that the generated scene remains true to the intended narrative atmosphere, aiding both general users and professional designers in visualizing the script's core themes. Unlike indoor scenes, which do not require consideration of audience perspective, StageDesigner incorporates a Foreground Projection Module to account directly for the front-facing audience view, managing sightlines to prevent unwanted occlusions and ensuring key background elements remain visible. This feature enhances the visual harmony of the scene, ensuring that important background elements are unobstructed and contribute to the overall atmosphere.

Furthermore, we introduce the \textbf{StagePro-V1} Dataset, the first dataset tailored specifically for artistic stage generation tasks. Unlike traditional 3D indoor scene datasets~\cite{fu20213d,scannet++}, which primarily contain standard furniture items with limited narrative context, the StagePro-V1 Dataset is curated by artists to capture the unique demands of theatrical scenography. Created in collaboration with professional stage designers, this dataset spans 276 real artistic stage productions from the 1940s to the 2020s and includes stage images, scripts (up to 1380 words), and 3D layouts that integrate spatial and thematic cues.
This resource allows for both quantitative and qualitative assessments of StageDesigner’s outputs, with benchmarks against adapted methods like LayoutGPT\cite{feng2024layoutgpt}, and user studies involving professional designers to validate the model’s practical relevance and alignment with theatrical design principles.

Our contributions can be summarized as follows:
\begin{itemize}
    \item We introduce StageDesigner, the first AI-driven framework specifically developed for automated artistic stage generation based on theater scripts.    
    \item We present the StagePro-V1, a comprehensive dataset for evaluating stage generation tasks, featuring annotated stage images, scripts, and 3D scene layouts.
    \item Extensive experiments and user studies, including both general users and professional stage designers, demonstrate the effectiveness of StageDesigner, with a 70\% higher favorability over LayoutGPT in both Layout Coherence and Overall Preference.

\end{itemize}

\section{Related Work}
\noindent \textbf{Indoor Scene Synthesis.} Several works have addressed indoor scene synthesis. ProcTHOR~\cite{deitke2022️} proposes a framework for the procedural generation of indoor scenes. Other works like~\cite{wang2021sceneformer,paschalidou2021atiss} generate indoor scenes by training transformers on a closed set of categories. More recent approaches leverage pretrained large language models (LLMs) to generate indoor scenes in a training-free manner, enabling open-vocabulary input. LayoutGPT~\cite{feng2024layoutgpt} focuses on generating a single indoor scene by using an LLM to arrange furniture within a CSS structure. Anyhome~\cite{fu2025anyhome} uses LLMs to produce floor plan bubble diagrams, which are then transformed into multi-room images via HouseGAN++~\cite{nauata2021house}. Holodeck~\cite{yang2024holodeck} proposes a refined subset of the Objaverse dataset~\cite{deitke2023objaverse} and employs LLMs to generate wall plans, creating multi-room environments suitable for embodied AI applications. SceneCraft~\cite{sc} achieves a coherent layout by continuously modifying a Python-based skill library. Unlike indoor scene synthesis, our work in artistic stage generation directly considers the front-facing audience view, managing sightlines to prevent unwanted occlusions and ensuring key elements remain visible.
\smallbreak
\noindent \textbf{Layout-Controlled Image Synthesis.} Applying layout control to place elements in specific user-defined locations within images has become an active research area. Previous works~\cite{li2021image,li2020bachgan,sun2019image,zhao2019image} used GAN-based methods to achieve layout control in image generation. Recently, approaches have focused on integrating layout control into Stable Diffusion~\cite{rombach2022high}. ReCo~\cite{yang2023reco} introduces a unified token vocabulary containing both text and positional tokens for precise, open-ended regional control. GLIGEN~\cite{li2023gligen} incorporates layout control through new trainable layers with a gated mechanism, while BoxDiff~\cite{xie2023boxdiff} implements layout control within cross-attention modules at each denoising timestep. Our work provides an ideal application for layout-controlled models, using precise regional control to generate unobstructed background images that complement foreground elements in immersive artistic stage scenes.
\smallbreak
\noindent \textbf{3D Scene Datasets.} Previous datasets for 3D scene synthesis focus largely on indoor environments and furniture setups~\cite{fu20213d,3d-future,scannet++,baruch2021arkitscenes,dai2017scannet,urbanscene3d}. While these datasets provide diverse furniture layouts and interior environments, they are limited to standard furniture items and lack the rich, narrative-driven descriptions necessary for stage design. Additionally, these datasets do not account for the unique perspective considerations required in theatrical scenography, where elements must remain visible from the audience’s fixed view. Our StagePro-V1 dataset addresses these limitations by incorporating detailed, script-based annotations specific to stage generation. Our dataset captures the complexity of theatrical environments, aligning spatial layouts with thematic cues from scripts, and serves as a foundational resource for AI-driven stage scenography research.

\section{Artistic-Stage Generation Formulation}

We define the artistic stage generation problem as the task of transforming an input theater script \( S \) into an immersive 3D stage environment that consists of two main components: a set of 3D foreground entities \( E \) with specific spatial layouts, and a 2D background \( B \) that aligns with the emotional tone and atmosphere of the script. 

Given a script \( S \), the goal is to generate:
\begin{itemize}
    \item  \textbf{Foreground layout} \( E = \{E_1, E_2, \dots, E_n\} \), where each \( E_i \) represents a 3D entity with properties such as position, size, orientation, and visual style.
    \item  \textbf{Background image} \( B \) that complements \( E \) by providing a coherent atmosphere in line with the script’s themes.
\end{itemize}

Mathematically, this can be represented as a function:
\[
G(S) \rightarrow (E, B),
\]
where \( G \) is a generative model that takes a script \( S \) as input and outputs the pair \( (E, B) \). Here:
\( E \) is designed to capture the essential physical elements described or implied in \( S \), avoiding spatial overlap and maintaining logical relationships between entities.
 \( B \) is generated to enhance the ambiance, ensuring that it does not obstruct critical foreground elements. This formulation allows the stage generation system to create visually compelling and thematically aligned 3D scenes based on script content.

\section{StageDesigner}
\begin{figure*}[!t]
\hsize=\textwidth
\centering
\includegraphics[width=\textwidth]{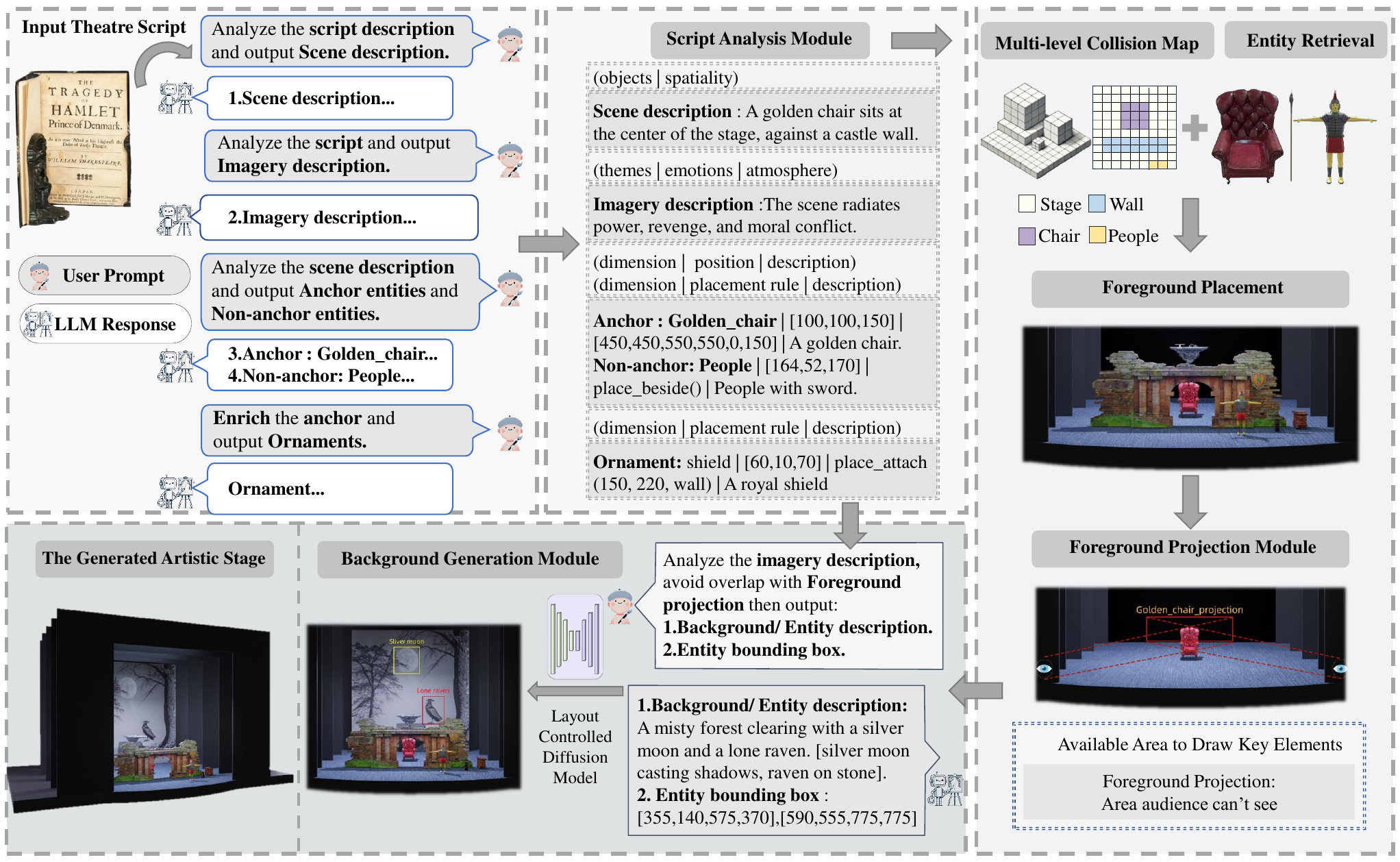}
    \caption{Overview of the StageDesigner pipeline. StageDesigner transforms an input theater script into a 3D stage layout through three main modules: (1) Script Analysis extracts key scene and imagery details; (2) Foreground Generation creates and places stage entities, retrieves corresponding 3D assets, and ensures spatial coherence using a multi-level collision map; (3) Background Generation produces a background image that complements the scene, guided by thematic elements and avoiding occlusions with foreground objects.}
    \label{fig: pipeline}
    \vspace{-0.5cm}
\end{figure*}
StageDesigner is a comprehensive system designed to transform an input theater script into a 3D foreground and 2D background that aligns with the intended scene's visual and thematic elements. The architecture of StageDesigner is divided into three main components: Script Analysis (Section~\ref{sec:script_module}), Foreground Generation (Section~\ref{sec:foreground}), and Background Generation (Section~\ref{sec:background}).
 
\smallbreak
\noindent \textbf{Overview.} The Script Analysis module processes the input script to extract key scene and imagery descriptions, setting the stage for subsequent modules.
The Foreground Generation includes the Entity Generation module, where foreground entities are generated based on the scene description, followed by the Multi-level Collision Map, which ensures that entities are placed without overlap. After generation, the Entity Retrieval Module retrieves matching 3D assets in the stage foreground. 
The Background Generation module begins with the Foreground Projection module, which calculates occlusion intervals for foreground entities to ensure that important background elements remain visible from the audience's perspective. Using the imagery description, the module creates a prompt and associates it with entity data and bounding boxes that avoid occlusion intervals, guiding layout control for background image generation. 
This results in a cohesive integration of the foreground and background, ensuring the final scene is visually coherent and faithful to the script’s thematic essence.

\subsection{Scripts Analysis}
\label{sec:script_module}
\noindent \textbf{Scripts Analysis Module.} This module is the first stage of StageDesigner, as shown in Table~\ref{tab:script_analysis}. The input script is processed and decomposed into two key components: Scene Description and Imagery Description. This decomposition removes noise from the raw script, allowing the model to focus on relevant details for subsequent modules. The Scene Description extracts entities and their spatial relationships, capturing visual elements, physical spaces, and layouts directly mentioned in the script to guide foreground generation. The Imagery Description captures the overarching themes, emotional tone, atmosphere, and symbolic meanings of the script, distilling the intended ambiance by removing extraneous details. This refined information guides the Background Generation Module to create a background that complements and enhances the stage atmosphere.
\begin{table}[ht]
  \centering
  \small
  \itshape
  \setlength{\abovecaptionskip}{1mm}
  \begin{tabular}{@{}p{1.2cm} p{6.3cm}@{}}
    \toprule
    Name & Description  \\
    \midrule
    Script & The stage is silent. The golden chair glistens under focused light at the center, with a mottled wall behind it casting dark, heavy shadows. Claudius stands beside the chair, one hand gripping its arm, tense. Hamlet steps forward from the shadows along the wall, his face a mix of anger and sorrow, seeming to blend with the depth of the shadows.\\ \hline
    Scene & \textbf{In the center of the stage} is \textbf{a golden chair}, symbolizing the power center of the royal palace. \textbf{The mottled wall and the interplay of light} and shadow create a historical ambiance. The lighting design focuses \textbf{on the center} from both sides. \\ \hline
    Imagery & The scene conveys a \underline{tense and mysterious} atmosphere, filled with \underline{themes of power}, \underline{revenge}, and \underline{moral struggle.}The overall mood is one of \underline{solemnity and intrigue}, reflecting the emotional weight of the unfolding power struggle.\\
    \bottomrule
  \end{tabular}
    \caption{Example of the Script Analysis Module, decomposing a raw script into Scene and Imagery Descriptions. \textbf{Bold} highlights entities and positions, while \underline{underlined} indicates atmosphere.}
  \vspace{-1em}
  \label{tab:script_analysis}
\end{table}

\subsection{Foreground Generation}
\label{sec:foreground}
\noindent \textbf{Entity Generation Module.} The Entity Generation Module shown in figure~\ref{fig: pipeline} processes the scene description to generate entities for the stage layout, consisting of Anchor Entity Generation and Ornament Generation. In the Anchor Entity Generation phase, anchor entities (e.g., a table) and their associated non-anchor entities (e.g., chairs) are generated. Each anchor entity's placement is defined using corner coordinates \([x_0, y_0]\) for the top-left corner and \([x_1, y_1]\) for the bottom-right corner, along with a height range \([h_0, h_1]\). Unlike single-point representations~\cite{feng2024layoutgpt}, corner coordinates clearly specify entity boundaries, ensuring that LLM-generated entities stay within the stage boundary. To enrich the scene, additional ornaments distinct from anchor groups are generated, adding variety to entity types. Each non-anchor entity and ornament includes dimensional information \([ \text{length}, \text{width}, \text{height} ]\), with precise placement managed by the Multi-level Collision Map module to prevent overlaps. Descriptions detailing materials, colors, textures, and design styles are also generated for all entities and used in the retrieval module to guide asset selection.
\smallbreak
\begin{figure}[!t]
\centering
\includegraphics[width=8.3cm]{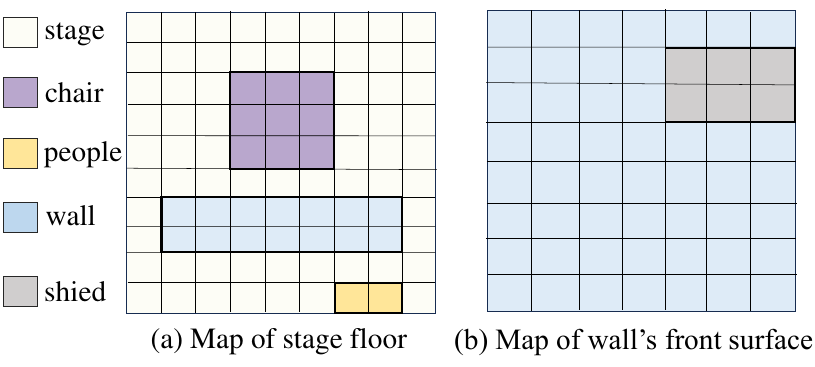}
    \caption{The example of collision map. (a) Collision map of stage floor. (b) Collision map of wall’s front surface.}
    \label{fig:collision}
    \vspace{-0.5cm}
\end{figure}
\smallbreak
\noindent \textbf{Multi-level Collision Map.} To ensure that entities are placed in reasonable, non-overlapping positions on the stage, we introduce the multi-level collision map shown in figure~\ref{fig:collision}. For an \( x \)-\( y \) stage floor of size \([N, N]\), we initialize an \( N \times N \) collision map, marking occupied positions as 1 and unoccupied positions as 0. For each anchor entity with dimensions \([L, W, H]\), we define collision maps for its front [\( L \times H \)], left [\( W \times H \)], right [\( W \times H \)], and top [\( L \times W \)] surfaces. Using this setup, we apply placement rules to accommodate various spatial relationships. For entities placed directly on the stage floor, such as objects positioned adjacent to the anchor, we search for an unoccupied position on the stage floor’s collision map near the anchor. For entities attached to surfaces, such as a painting on a wall, we search for an all-zero region on the anchor’s front, left, or right surfaces at a height \( h \) generated by the LLM. Finally, for entities placed on top of another, like a teacup on a table, we locate an unoccupied area on the entity’s top surface. 
\smallbreak
\noindent \textbf{Entities Retrieval Module.} To populate the foreground, we retrieve entities from a subset of the Objaverse dataset~\cite{deitke2023objaverse} introduced in the Holodeck~\cite{yang2024holodeck}. For each entity, we use the generated entity name and description, which are concatenated to form the input text. Similarity is computed by embedding this concatenated text with dataset images using CLIP, while Sentence-BERT is used to compare the concatenated text with dataset entity names. We combine the two scores, then randomly select one entity from the top 10 highest-scoring matches above a threshold.

\subsection{Background Generation}
\label{sec:background}
\noindent \textbf{Foreground Projection Module.}
To ensure that 3D foreground entities do not obstruct background elements from the audience's perspective. We model the audience's view as parallel rays and define the coordinates of the leftmost and rightmost front-row viewers, as shown in the right-bottom of figure~\ref{fig: pipeline}. Each entity’s projection onto the background is determined by tracing sightlines from leftmost and rightmost viewer positions along the entity's left and right edges. The resulting bounding boxes encompass all regions obscured from the view of audience seated between the leftmost and rightmost. This method preserves the visibility of important background elements, ensuring that the scene maintains its visual and thematic integrity.
\smallbreak
\noindent \textbf{Background Generation Module.} This module, shown at the bottom of Figure~\ref{fig: pipeline}, creates a cohesive background that aligns with the script’s atmosphere and integrates seamlessly with the foreground entities. Using the imagery description from the Script Analysis Module and the bounding boxes from the Foreground Projection Module, we generate the background prompt that reflects the thematic and emotional tone. This prompt includes entities positioned within specified bounding boxes that avoid overlapping with the foreground projection, ensuring key background elements remain visible. The prompt is then input into a layout-controlled diffusion model to produce the final background.

\section{StagePro-V1 Dataset}
The StagePro-v1 dataset is a comprehensive resource for AI-driven stage generation, created in collaboration with professional stage designers to address the lack of datasets tailored for scenography. Compiling a total of 276 unique stage models from productions spanning the 1940s to the 2020s, the dataset represents a wide range of styles. Among these, 271 stages are categorized as Proscenium, 267 as Drama, 150 as Neo-Realism, and 47 as Symbolism. Some stages are classified under multiple stylistic categories, providing a rich diversity of design elements across the dataset. It is organized into three components: images, scripts, and layouts, providing a rich resource for diverse scenographic contexts. Dataset construction involved capturing both frontal and top-down images of each artistic stage. 
\begin{figure}[!t]
\centering
\includegraphics[width=8.3cm]{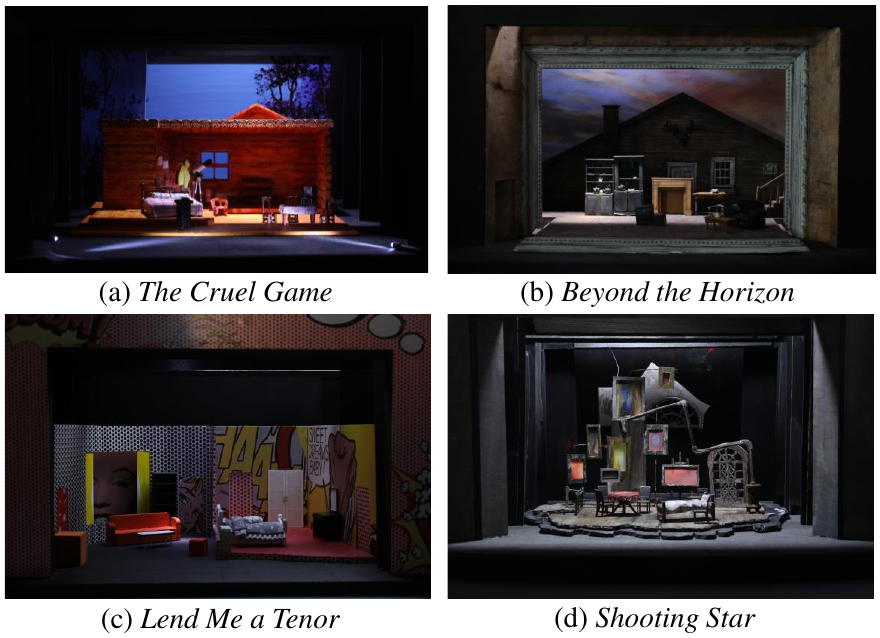}
    \caption{The stage samples in our dataset. (a) The Cruel Game, (b) Beyond the Horizon,
(c) Lend Me a Tenor, (d) Shooting Star. }
    \label{fig:dataset}
    \vspace{-0.5cm}
\end{figure}
\smallbreak
\noindent \textbf{For the scripts}, we sourced complete books corresponding to each stage and matched them with their respective stage images. Given the substantial length of each stage's corresponding book, Ernie Bot~\cite{sun2019ernie} was used to analyze these books and extract essential information, such as main themes, story settings, and scene locations. When full books were unavailable, we prompted Ernie Bot with the title of the stage's book to gather foundational insights. Once the core information was established, images were provided to Ernie Bot with specific queries, such as: \textit{Based on the main themes, story background, and setting of the play, describe the stage script depicted in this image, including the overall design style, layout, main props, entity, and their spatial relationships. Additionally, describe the entity styles.} This process allowed Ernie Bot to draft detailed stage scene scripts. Those initial scripts were refined by stage design artists to ensure accuracy, removing redundancies, correcting inaccuracies, and adding missing details. The scripts in the dataset range from 56 to 1380 words, covering both concise and highly detailed descriptions. 
\smallbreak
\noindent \textbf{For the layouts}, entity heights were annotated from frontal views, while top-down views provided \( xy \)-plane positions, orientations, and categories of the entities. The position data was stored as \texttt{left:}\([x_0, y_0]\), \texttt{right:}\([x_1, y_1]\), \texttt{h:}\([h_0, h_1]\), where \texttt{left} and \texttt{right} represent the corner coordinates, and \texttt{h} denotes the lowest and highest points of the entity, ensuring a consistent spatial representation. Entity counts per stage range from 1 to 21, resulting in a dataset that captures intricate visual and spatial details.

\section{Experiments}
\begin{figure*}[!t]
\hsize=\textwidth
\centering
\includegraphics[width=\textwidth]{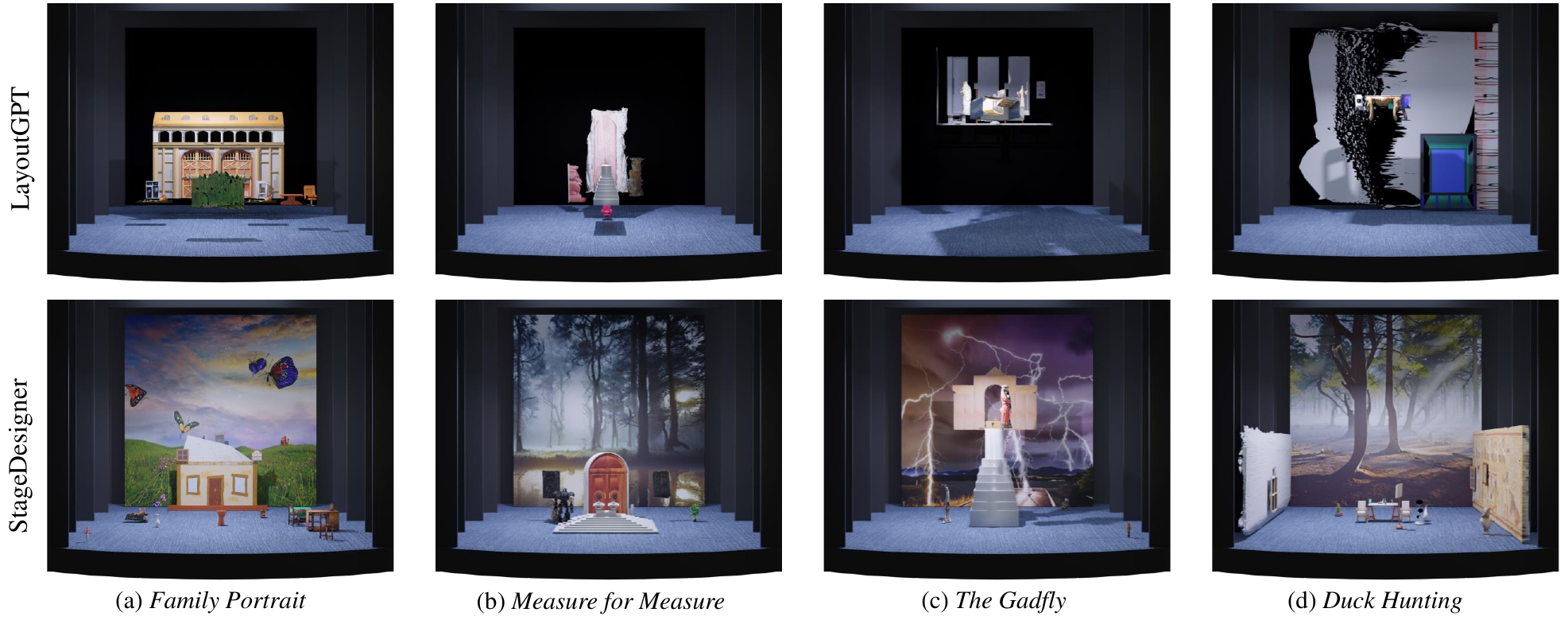}
    \caption{Qualitative comparison between LayoutGPT and StageDesigner. The first row shows stages generated by LayoutGPT, and the second row shows stages generated by StageDesigner. StageDesigner results show better entity placement and atmospheric expression. e.g.``~\textit{Family Portrait} '' denotes the theater script.}
    \label{fig: qualitative}
    \vspace{-0.5cm}
\end{figure*}

\subsection{Experiment Setups}

\noindent \textbf{Datasets.}  Since StageDesigner is a training-free system, all the 276 stages from the newly introduced StagePro-V1 dataset were used exclusively for testing. This dataset provided a comprehensive basis to validate the system’s ability to generate coherent and thematically appropriate stage designs. For the entity retrieval component, we utilized a subset of the Objaverse~\cite{deitke2023objaverse} dataset, which consists of 50,092 assets annotated by Holodeck~\cite{yang2024holodeck}. This subset ensured a diverse and high-quality pool of 3D models suitable for accurately populating the stage foreground.
\smallbreak
\noindent \textbf{Evaluation Metrics.} we use a combination of established metrics and new metrics defined specifically for this work.
Established Metrics: \textit{1.Out-of-Bound (OOB).} This metric calculates the average volume of foreground entities that extend beyond the stage floor, reflecting the model’s ability to respect stage boundaries. \textit{2.CLIP Similarity (CLIP-sim).} CLIP-sim computes the similarity between dataset scripts and the final rendered stage images, indicating how well the generated stage aligns with the script’s thematic and visual elements.
Newly Defined Metrics: \textit{3.Overlap-inter-Stage (OIS).} This metric calculates the average overlapping volume between foreground entities within each stage, assessing the model’s ability to generate a spatially coherent layout. \textit{4.Intersection-with-Ground truth (IWG).} IWG calculates the average intersection volume between generated layouts and the ground truth layouts of foreground entities, providing a measure of layout accuracy. \textit{5.Class Diversity.} This metric calculates the average number of unique entity categories generated per stage, indicating the model’s ability to produce a diverse range of entities.

\smallbreak
\noindent \textbf{Baselines.} Since our work is the first to address stage generation, there are no existing methods for direct comparison. To establish a baseline, we adapted LayoutGPT~\cite{feng2024layoutgpt}, a model originally designed to generate CSS-based entity layouts for single indoor scenes using LLMs. We modified LayoutGPT to accept scripts as input and generate corresponding stage entity layouts. For a fair comparison, LayoutGPT was also given access to the subset of Objaverse used in our method. 
\smallbreak
\noindent \textbf{Implementation Details.} The stage size in our paper is simply defined as \(1000 \times 1000 \times 1000 \ cm^3\). Both StageDesigner and the adapted LayoutGPT baseline use OpenAI's gpt-4o with a temperature of 0.7, a maximum token limit of 2048, and top-p set to 1.0 as the LLM model. In StageDesigner, the Layout Controlled Diffusion Model is based on the ReCo~\cite{yang2023reco} model, trained on the LAION-2b~\cite{schuhmann2022laion} dataset. For entity retrieval, StageDesigner uses OpenCLIP~\cite{ilharco_gabriel_2021_5143773} (ViT-L-14 variant trained on the LAION-2b dataset) combined with Sentence-BERT~\cite{reimers2019sentence} for text similarity matching, with a retrieval threshold of 27. The same OpenCLIP model is used for the CLIP-sim metric.
\subsection{Experiment Results} 
\subsubsection{Quantitative Analysis}
\noindent \textbf{Foreground Layout Coherence.} Table~\ref{tab:Layout experiment} presents a quantitative comparison between StageDesigner and the adapted LayoutGPT baseline using key stage generation metrics. 

For the Out-of-Bound metric, StageDesigner demonstrates a significantly lower average out-of-bound volume (0.0468 \(m^{3}\)) compared to LayoutGPT (6.46 \(m^{3}\)). This can be attributed to StageDesigner’s method of letting the LLM predict coordinates only for anchor entities and using corner coordinate representation for placement. This approach allows the LLM to clearly understand the boundaries of each entity, making it easier to avoid out-of-bounds placements. In contrast, LayoutGPT generates coordinates based on a single-point position with length, width, and height, which makes direct boundary checks more challenging and results in a higher likelihood of out-of-bounds errors.

For the Overlap-Inter-Stage (OIS) metric, StageDesigner also shows superior performance with a lower overlap volume (0.756 \(m^{3}\)) compared to LayoutGPT (18.2 \(m^{3}\)). This is due to StageDesigner’s strategy of limiting direct coordinate prediction by the LLM to only a few anchor entities, which reduces the potential for overlaps between entities. Additionally, the Multi-level Collision Map ensures that non-anchor entities and ornaments are positioned in vacant areas of the stage floor or on available surfaces of other entities. This systematic search helps maintain spatial coherence and minimizes overlap within the generated layout.

For the IWG metric, LayoutGPT scores higher (14.5 \(m^{3}\)) than StageDesigner (9.03 \(m^{3}\)). However, this higher score can be attributed to LayoutGPT’s tendency to generate entities with dimensions that are too large, leading to greater overlap with ground truth entities but at the cost of more frequent out-of-bounds, unrealistic and overlapping issues.
\smallbreak
\noindent \textbf{Diversity and Thematic Alignment.} Table~\ref{tab: overall experiment} shows the comparison based on Class Diversity and CLIP Similarity (CLIP-sim). StageDesigner achieves a higher average class diversity (11.7) compared to LayoutGPT (7.46), indicating its capability to create varied and complex stage layouts. The Min/Max Classes metric supports this, with StageDesigner generating between 5 and 22 unique classes per stage, while LayoutGPT produces a narrower range of 2 to 18. This demonstrates StageDesigner’s enhanced ability to reflect a richer array of entities, leading to more realistic and comprehensive stage designs. In terms of CLIP-sim, StageDesigner scores higher (30.3) than LayoutGPT (29.1), showcasing that the stage designs produced by StageDesigner are better aligned with the thematic and visual cues present in the input scripts.

\begin{table}
    \centering
    \resizebox{\linewidth}{!}{
    \begin{tabular}{lccc}
    \toprule
    Method   &Out-of-Bound(\(m^{3}\)) $\downarrow$&OIS(\(m^{3}\)) $\downarrow$ &IWG(\(m^{3}\)) $\uparrow$  \\
    \midrule
    LayoutGPT$^*$~\cite{feng2024layoutgpt}     &6.46 &18.2 &14.5  \\
    StageDesigner                          &0.0468 &0.756 &9.03  \\
    \bottomrule
    \end{tabular}}
    \caption{Quantitative Comparison of Methods Using Stage Metrics for Spatial Accuracy and Coherence. ``$*$'' denote the results is conducted by our experiment.}
    \label{tab:Layout experiment}
    \vspace{-1em}
\end{table}

\begin{table}
    \centering
    \resizebox{\linewidth}{!}{
    \begin{tabular}{lccc}
    \toprule
    Method   &Class Diversity $\uparrow$&CLIP-sim $\uparrow$ &Min/Max Classes \\
    \midrule
    LayoutGPT$^*$~\cite{feng2024layoutgpt}      &7.46 &29.1&2/18 \\
    StageDesigner                           &11.7 &30.3&5/22  \\
    \bottomrule
    \end{tabular}}
    \caption{Comparison of Methods Based on Class Diversity, CLIP Similarity, and Min/Max generated classes in single stage. ``$*$'' denote the results is conducted by our experiment.}
    \label{tab: overall experiment}
    \vspace{-1em}
\end{table}

\smallbreak
\subsubsection{Qualitative Analysis}
\noindent \textbf{Qualitative Results and Analysis.} Figure~\ref{fig: qualitative} compares the stage designs generated by StageDesigner and LayoutGPT. StageDesigner outputs show rich foreground entities, well-organized layouts, and backgrounds that enhance the stage atmosphere while carefully avoiding overlaps with key foreground elements, such as the butterflies in the \textit{Family Portrait} scene. This demonstrates StageDesigner’s strength in generating cohesive and thematic stage environments with realistic spatial arrangements. In contrast, LayoutGPT-generated stages feature fewer entity types and often display layout inconsistencies, including mispositioned, floating entities due to incorrect placement predictions. Additionally, LayoutGPT does not have a background generation module, so its stage designs lack contextual and atmospheric elements, making the scenes less immersive.
\begin{table}
    \centering
    \resizebox{\linewidth}{!}{
    \begin{tabular}{lcc}
    \toprule
    Metric   &StageDesigner &StageDesigner w/o Projection \\
    \midrule
     Overlap \(\downarrow\)    &2.70 & 17.7\\
    \bottomrule
    \end{tabular}}
   \caption{Overlap comparison between background elements and foreground projections in StageDesigner, with and without the Foreground Projection module.}
    \label{tab: ablation_overlap}
    \vspace{-1em}
\end{table}

\begin{table}
    \centering
    \resizebox{\linewidth}{!}{
    \begin{tabular}{lcc}
    \toprule
    Metric   &StageDesigner &StageDesigner w/o background \\
    \midrule
     CLIP-sim \(\uparrow\)     &30.3 &32.1 \\
    \bottomrule
    \end{tabular}}
    \caption{Comparison of CLIP Similarity (CLIP-sim) between StageDesigner with and without background images.}
    \label{tab: ablation_clip}
    \vspace{-1em}
\end{table}
\smallbreak
\subsubsection{User Study}
\noindent \textbf{Comparative User Study.}  We conducted a user study comparing StageDesigner and LayoutGPT on two aspects: (1) \textit{Layout Coherence}: Evaluated the realism and appropriateness of foreground entity categories, layout, and spatial relationships. (2) \textit{Overall Preference}: Considered the entire stage scene and atmosphere, focusing on which design participants found more visually appealing. Using 10 scripts, we generated corresponding stage designs with both StageDesigner and LayoutGPT. We then invited 69 general users and 14 stage design experts to evaluate the designs, with results displayed in Figure~\ref{fig:userstudy} (a) and (b) (smiley icons mark expert responses).
Overall, StageDesigner was preferred over LayoutGPT on both metrics. Users consistently favored StageDesigner for its realistic layouts and cohesive atmospheres, demonstrating its ability to capture functional and aesthetic aspects of stage design effectively.

\subsection{Ablation Study}
\subsubsection{Quantitative Analysis}

\noindent \textbf{Foreground Projection.}
Table~\ref{tab: ablation_overlap} shows a significant reduction in overlap between background elements generated by LLM and foreground projections when the Foreground Projection module is applied, from 17.7 to 2.70. This result demonstrates the module's effectiveness in preserving background visibility and maintaining visual coherence. Without the module, the increased overlap leads to foreground entities frequently obstructing key background elements, reducing the overall thematic and visual quality of the scene.

\smallbreak
\noindent \textbf{CLIP-sim Analysis on Background.} Table~\ref{tab: ablation_clip} shows that StageDesigner without background images has a slightly higher CLIP-sim score than with backgrounds. This difference reveals two main limitations of CLIP when evaluating complex scripts. First, CLIP’s 77-token limit leads to truncation of longer scripts (up to 1380 words in our dataset). As a result, we are compelled to truncate the entire script for encoding, summing the resulting vectors. This limitation reduces CLIP’s ability to accurately capture the script’s thematic content. Second, CLIP primarily identifies specific object-image matches rather than overall atmosphere. Background elements, while enhancing the scene’s ambiance, may introduce “noise” that lowers the similarity score by shifting focus away from foreground objects.

\begin{figure}[!t]
\centering
\includegraphics[width=8.3cm]{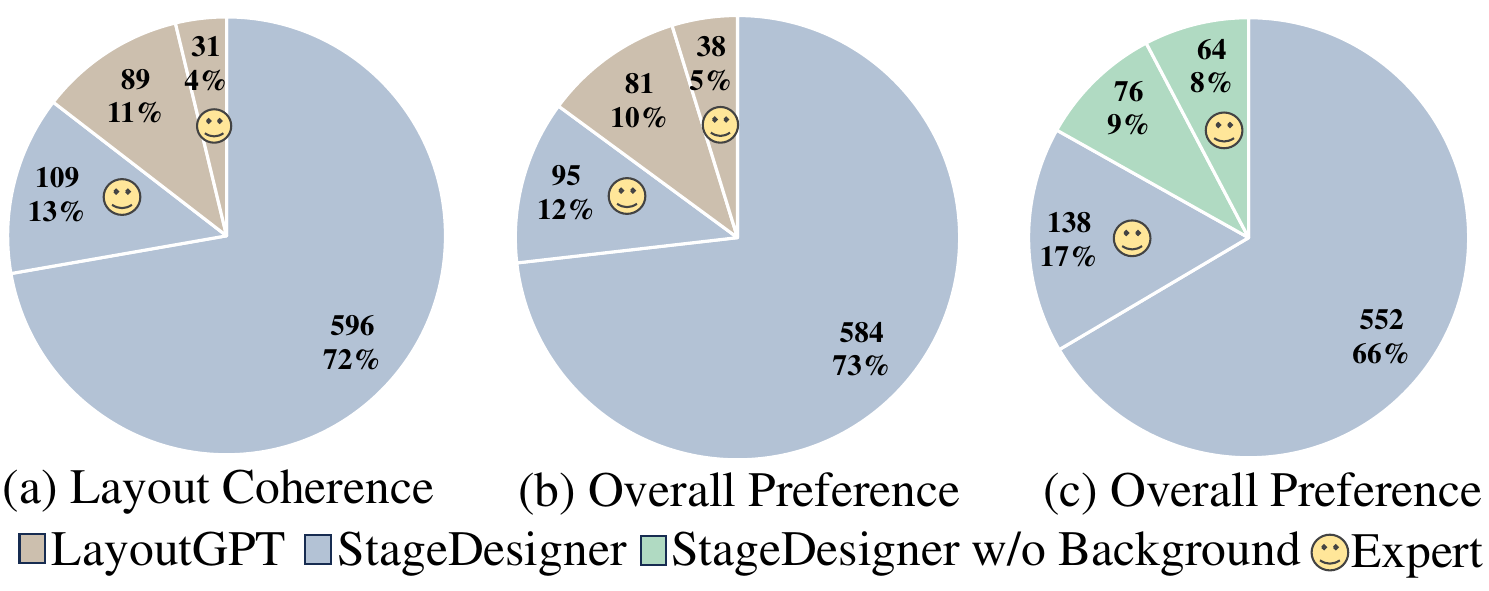}
    \caption{User study: (a) Layout Coherence comparison with LayoutGPT. (b) Overall Preference comparison with LayoutGPT. (c) Ablation study on Overall Preference for StageDesigner with and without background. Smiley icons mark expert responses.}
    \label{fig:userstudy}
    \vspace{-1em}
\end{figure}
\begin{figure}[!t]
\centering
\includegraphics[width=8.3cm]{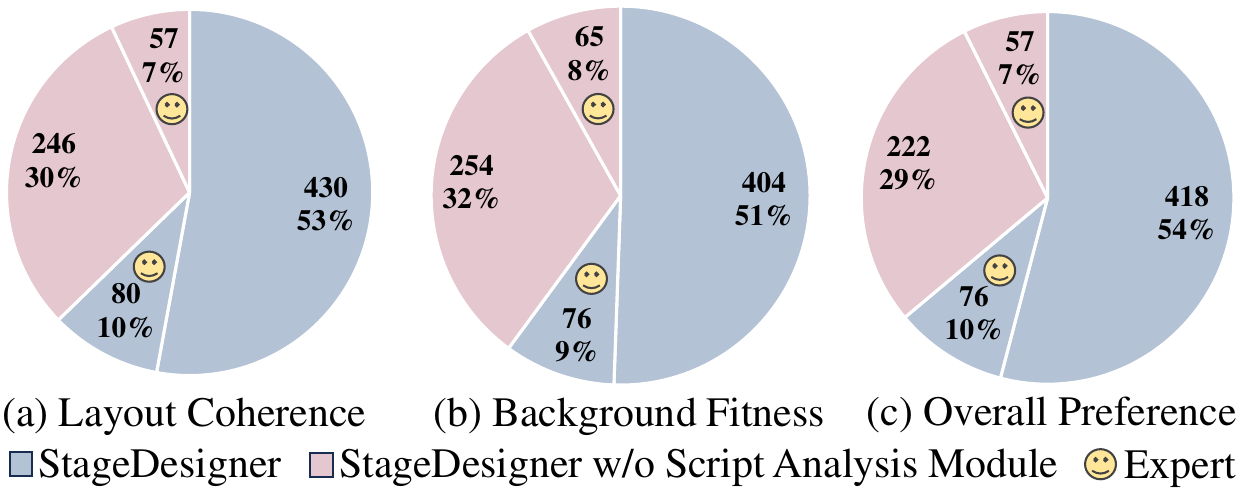}
    \caption{Ablation study on StageDesigner's Script Analysis Module. (a) Layout Coherence result. (b) Background Fitness result. (c) Overall Preference result. Smiley icons mark expert responses.}
    \label{fig:userstudy1}
    \vspace{-0.5cm}
\end{figure}
\subsubsection{User Study}
\noindent \textbf{Impact of Background.}
To further evaluate the importance of background elements, we conducted an ablation study using StageDesigner-generated stages. We invited the same 69 general users and 14 stage design experts from the previous study. Using 10 scripts, we generated two versions of each stage: one with a background and one without, keeping the foreground identical. Participants selected their preferred design based on \textit{Overall Preference}. As shown in Figure~\ref{fig:userstudy} (c), the results reveal a strong preference for designs with backgrounds, with significantly more participants favoring these over the background-free versions. This suggests that background elements play a crucial role in enhancing stage atmosphere and thematic depth, creating a more immersive and cohesive experience by reinforcing the context and tone of the scene.
\smallbreak
\noindent \textbf{Impact of Script Analysis Module.}
To evaluate the impact of the Script Analysis Module on stage design quality, we compared StageDesigner with and without this module. We invited the same 69 general users and 14 stage design experts as in previous studies. Using 10 scripts different from those previously used, we generated two versions of each stage: one with the Script Analysis Module and one without. Participants evaluated each version on three criteria: (1) \textit{Layout Coherence}: as previously defined. (2) \textit{Overall Preference}: as previously defined. (3) \textit{Background Fitness}: Harmony between the generated background and the intended stage atmosphere. As shown in Figure~\ref{fig:userstudy1}, results indicate that StageDesigner with the Script Analysis Module scored higher across all three criteria. These findings underscore the Script Analysis Module’s role in enhancing the thematic alignment and coherence of both layout and background, resulting in a more visually compelling scene.

\section{Conclusion} 
 We introduced StageDesigner, an AI-based framework that transforms scripts into cohesive 3D stage layouts using LLMs and layout-controlled diffusion models. By structuring the generation process into Script Analysis, Foreground Generation, and Background Generation, StageDesigner ensures spatial coherence and thematic alignment. Our StagePro-V1 dataset, curated with professional input, serves as a foundational resource for stage design research. 
\section{Acknowledgements}
This work is supported by 2024 CCF-BAIDU OPEN FUND (CCF-BAIDU OF202424), the National Natural Science Foundation of China (No.62402306) and the Natural Science Foundation of Shanghai (No.24ZR1422400).

{
    \small
    \bibliographystyle{ieeenat_fullname}
    \bibliography{main}

\begin{thebibliography}{31}
\providecommand{\natexlab}[1]{#1}
\providecommand{\url}[1]{\texttt{#1}}
\expandafter\ifx\csname urlstyle\endcsname\relax
  \providecommand{\doi}[1]{doi: #1}\else
  \providecommand{\doi}{doi: \begingroup \urlstyle{rm}\Url}\fi

\bibitem[Baruch et~al.(2021)Baruch, Chen, Dehghan, Dimry, Feigin, Fu, Gebauer, Joffe, Kurz, Schwartz, et~al.]{baruch2021arkitscenes}
Gilad Baruch, Zhuoyuan Chen, Afshin Dehghan, Tal Dimry, Yuri Feigin, Peter Fu, Thomas Gebauer, Brandon Joffe, Daniel Kurz, Arik Schwartz, et~al.
\newblock Arkitscenes: A diverse real-world dataset for 3d indoor scene understanding using mobile rgb-d data.
\newblock \emph{arXiv preprint arXiv:2111.08897}, 2021.

\bibitem[Dai et~al.(2017)Dai, Chang, Savva, Halber, Funkhouser, and Nie{\ss}ner]{dai2017scannet}
Angela Dai, Angel~X Chang, Manolis Savva, Maciej Halber, Thomas Funkhouser, and Matthias Nie{\ss}ner.
\newblock Scannet: Richly-annotated 3d reconstructions of indoor scenes.
\newblock In \emph{Proceedings of the IEEE conference on computer vision and pattern recognition}, pages 5828--5839, 2017.

\bibitem[Deitke et~al.(2022)Deitke, VanderBilt, Herrasti, Weihs, Ehsani, Salvador, Han, Kolve, Kembhavi, and Mottaghi]{deitke2022️}
Matt Deitke, Eli VanderBilt, Alvaro Herrasti, Luca Weihs, Kiana Ehsani, Jordi Salvador, Winson Han, Eric Kolve, Aniruddha Kembhavi, and Roozbeh Mottaghi.
\newblock Procthor: Large-scale embodied ai using procedural generation.
\newblock \emph{Advances in Neural Information Processing Systems}, 35:\penalty0 5982--5994, 2022.

\bibitem[Deitke et~al.(2023)Deitke, Schwenk, Salvador, Weihs, Michel, VanderBilt, Schmidt, Ehsani, Kembhavi, and Farhadi]{deitke2023objaverse}
Matt Deitke, Dustin Schwenk, Jordi Salvador, Luca Weihs, Oscar Michel, Eli VanderBilt, Ludwig Schmidt, Kiana Ehsani, Aniruddha Kembhavi, and Ali Farhadi.
\newblock Objaverse: A universe of annotated 3d objects.
\newblock In \emph{Proceedings of the IEEE/CVF Conference on Computer Vision and Pattern Recognition}, pages 13142--13153, 2023.

\bibitem[Essin(2012)]{essin2012stage}
E Essin.
\newblock \emph{Stage Designers in Early Twentieth-Century America: Artists, Activists, Cultural Critics}.
\newblock Springer, 2012.

\bibitem[Feng et~al.(2024)Feng, Zhu, Fu, Jampani, Akula, He, Basu, Wang, and Wang]{feng2024layoutgpt}
Weixi Feng, Wanrong Zhu, Tsu-jui Fu, Varun Jampani, Arjun Akula, Xuehai He, Sugato Basu, Xin~Eric Wang, and William~Yang Wang.
\newblock Layoutgpt: Compositional visual planning and generation with large language models.
\newblock \emph{Advances in Neural Information Processing Systems}, 36, 2024.

\bibitem[Fu et~al.(2021{\natexlab{a}})Fu, Cai, Gao, Zhang, Wang, Li, Zeng, Sun, Jia, Zhao, et~al.]{fu20213d}
Huan Fu, Bowen Cai, Lin Gao, Ling-Xiao Zhang, Jiaming Wang, Cao Li, Qixun Zeng, Chengyue Sun, Rongfei Jia, Binqiang Zhao, et~al.
\newblock 3d-front: 3d furnished rooms with layouts and semantics.
\newblock In \emph{Proceedings of the IEEE/CVF International Conference on Computer Vision}, pages 10933--10942, 2021{\natexlab{a}}.

\bibitem[Fu et~al.(2021{\natexlab{b}})Fu, Jia, Gao, Gong, Zhao, Maybank, and Tao]{3d-future}
Huan Fu, Rongfei Jia, Lin Gao, Mingming Gong, Binqiang Zhao, Steve Maybank, and Dacheng Tao.
\newblock 3d-future: 3d furniture shape with texture.
\newblock \emph{International Journal of Computer Vision}, 129:\penalty0 3313--3337, 2021{\natexlab{b}}.

\bibitem[Fu et~al.(2025)Fu, Wen, Liu, and Sridhar]{fu2025anyhome}
Rao Fu, Zehao Wen, Zichen Liu, and Srinath Sridhar.
\newblock Anyhome: Open-vocabulary generation of structured and textured 3d homes.
\newblock In \emph{European Conference on Computer Vision}, pages 52--70. Springer, 2025.

\bibitem[Hu et~al.(2024)Hu, Iscen, Jain, Kipf, Yue, Ross, Schmid, and Fathi]{sc}
Ziniu Hu, Ahmet Iscen, Aashi Jain, Thomas Kipf, Yisong Yue, David~A Ross, Cordelia Schmid, and Alireza Fathi.
\newblock Scenecraft: An llm agent for synthesizing 3d scenes as blender code.
\newblock In \emph{Forty-first International Conference on Machine Learning}, 2024.

\bibitem[Ilharco et~al.(2021)Ilharco, Wortsman, Wightman, Gordon, Carlini, Taori, Dave, Shankar, Namkoong, Miller, Hajishirzi, Farhadi, and Schmidt]{ilharco_gabriel_2021_5143773}
Gabriel Ilharco, Mitchell Wortsman, Ross Wightman, Cade Gordon, Nicholas Carlini, Rohan Taori, Achal Dave, Vaishaal Shankar, Hongseok Namkoong, John Miller, Hannaneh Hajishirzi, Ali Farhadi, and Ludwig Schmidt.
\newblock Openclip, 2021.
\newblock If you use this software, please cite it as below.

\bibitem[Li et~al.(2020)Li, Cheng, Gan, Yu, Wang, and Liu]{li2020bachgan}
Yandong Li, Yu Cheng, Zhe Gan, Licheng Yu, Liqiang Wang, and Jingjing Liu.
\newblock Bachgan: High-resolution image synthesis from salient object layout.
\newblock In \emph{Proceedings of the IEEE/CVF Conference on Computer Vision and Pattern Recognition}, pages 8365--8374, 2020.

\bibitem[Li et~al.(2023)Li, Liu, Wu, Mu, Yang, Gao, Li, and Lee]{li2023gligen}
Yuheng Li, Haotian Liu, Qingyang Wu, Fangzhou Mu, Jianwei Yang, Jianfeng Gao, Chunyuan Li, and Yong~Jae Lee.
\newblock Gligen: Open-set grounded text-to-image generation.
\newblock In \emph{Proceedings of the IEEE/CVF Conference on Computer Vision and Pattern Recognition}, pages 22511--22521, 2023.

\bibitem[Li et~al.(2021)Li, Wu, Koh, Tang, and Sun]{li2021image}
Zejian Li, Jingyu Wu, Immanuel Koh, Yongchuan Tang, and Lingyun Sun.
\newblock Image synthesis from layout with locality-aware mask adaption.
\newblock In \emph{Proceedings of the IEEE/CVF International Conference on Computer Vision}, pages 13819--13828, 2021.

\bibitem[Lin et~al.(2022)Lin, Liu, Hu, Yan, Xie, and Huang]{urbanscene3d}
Liqiang Lin, Yilin Liu, Yue Hu, Xingguang Yan, Ke Xie, and Hui Huang.
\newblock Capturing, reconstructing, and simulating: the urbanscene3d dataset.
\newblock In \emph{European Conference on Computer Vision}, pages 93--109. Springer, 2022.

\bibitem[Nauata et~al.(2021)Nauata, Hosseini, Chang, Chu, Cheng, and Furukawa]{nauata2021house}
Nelson Nauata, Sepidehsadat Hosseini, Kai-Hung Chang, Hang Chu, Chin-Yi Cheng, and Yasutaka Furukawa.
\newblock House-gan++: Generative adversarial layout refinement network towards intelligent computational agent for professional architects.
\newblock In \emph{Proceedings of the IEEE/CVF Conference on Computer Vision and Pattern Recognition}, pages 13632--13641, 2021.

\bibitem[Paschalidou et~al.(2021)Paschalidou, Kar, Shugrina, Kreis, Geiger, and Fidler]{paschalidou2021atiss}
Despoina Paschalidou, Amlan Kar, Maria Shugrina, Karsten Kreis, Andreas Geiger, and Sanja Fidler.
\newblock Atiss: Autoregressive transformers for indoor scene synthesis.
\newblock \emph{Advances in Neural Information Processing Systems}, 34:\penalty0 12013--12026, 2021.

\bibitem[Reimers(2019)]{reimers2019sentence}
N Reimers.
\newblock Sentence-bert: Sentence embeddings using siamese bert-networks.
\newblock \emph{arXiv preprint arXiv:1908.10084}, 2019.

\bibitem[Rombach et~al.(2022)Rombach, Blattmann, Lorenz, Esser, and Ommer]{rombach2022high}
Robin Rombach, Andreas Blattmann, Dominik Lorenz, Patrick Esser, and Bj{\"o}rn Ommer.
\newblock High-resolution image synthesis with latent diffusion models.
\newblock In \emph{Proceedings of the IEEE/CVF conference on computer vision and pattern recognition}, pages 10684--10695, 2022.

\bibitem[Schuhmann et~al.(2022)Schuhmann, Beaumont, Vencu, Gordon, Wightman, Cherti, Coombes, Katta, Mullis, Wortsman, et~al.]{schuhmann2022laion}
Christoph Schuhmann, Romain Beaumont, Richard Vencu, Cade Gordon, Ross Wightman, Mehdi Cherti, Theo Coombes, Aarush Katta, Clayton Mullis, Mitchell Wortsman, et~al.
\newblock Laion-5b: An open large-scale dataset for training next generation image-text models.
\newblock \emph{Advances in Neural Information Processing Systems}, 35:\penalty0 25278--25294, 2022.

\bibitem[Stern and Gold(2021)]{stern2021stage}
Lawrence Stern and Jill Gold.
\newblock \emph{Stage management}.
\newblock Routledge, 2021.

\bibitem[Sun and Wu(2019)]{sun2019image}
Wei Sun and Tianfu Wu.
\newblock Image synthesis from reconfigurable layout and style.
\newblock In \emph{Proceedings of the IEEE/CVF International Conference on Computer Vision}, pages 10531--10540, 2019.

\bibitem[Sun et~al.(2019)Sun, Wang, Li, Feng, Chen, Zhang, Tian, Zhu, Tian, and Wu]{sun2019ernie}
Yu Sun, Shuohuan Wang, Yukun Li, Shikun Feng, Xuyi Chen, Han Zhang, Xin Tian, Danxiang Zhu, Hao Tian, and Hua Wu.
\newblock Ernie: Enhanced representation through knowledge integration.
\newblock \emph{arXiv preprint arXiv:1904.09223}, 2019.

\bibitem[Wang et~al.(2021)Wang, Yeshwanth, and Nie{\ss}ner]{wang2021sceneformer}
Xinpeng Wang, Chandan Yeshwanth, and Matthias Nie{\ss}ner.
\newblock Sceneformer: Indoor scene generation with transformers.
\newblock In \emph{2021 International Conference on 3D Vision (3DV)}, pages 106--115. IEEE, 2021.

\bibitem[Wolf and Block(2014)]{wolf2014scene}
R~Craig Wolf and Dick Block.
\newblock \emph{Scene design and stage lighting}.
\newblock Wadsworth, Cengage Learning, 2014.

\bibitem[Xie et~al.(2023)Xie, Li, Huang, Liu, Zhang, Zheng, and Shou]{xie2023boxdiff}
Jinheng Xie, Yuexiang Li, Yawen Huang, Haozhe Liu, Wentian Zhang, Yefeng Zheng, and Mike~Zheng Shou.
\newblock Boxdiff: Text-to-image synthesis with training-free box-constrained diffusion.
\newblock In \emph{Proceedings of the IEEE/CVF International Conference on Computer Vision}, pages 7452--7461, 2023.

\bibitem[Yang et~al.(2024)Yang, Sun, Weihs, VanderBilt, Herrasti, Han, Wu, Haber, Krishna, Liu, et~al.]{yang2024holodeck}
Yue Yang, Fan-Yun Sun, Luca Weihs, Eli VanderBilt, Alvaro Herrasti, Winson Han, Jiajun Wu, Nick Haber, Ranjay Krishna, Lingjie Liu, et~al.
\newblock Holodeck: Language guided generation of 3d embodied ai environments.
\newblock In \emph{Proceedings of the IEEE/CVF Conference on Computer Vision and Pattern Recognition}, pages 16227--16237, 2024.

\bibitem[Yang et~al.(2023)Yang, Wang, Gan, Li, Lin, Wu, Duan, Liu, Liu, Zeng, et~al.]{yang2023reco}
Zhengyuan Yang, Jianfeng Wang, Zhe Gan, Linjie Li, Kevin Lin, Chenfei Wu, Nan Duan, Zicheng Liu, Ce Liu, Michael Zeng, et~al.
\newblock Reco: Region-controlled text-to-image generation.
\newblock In \emph{Proceedings of the IEEE/CVF Conference on Computer Vision and Pattern Recognition}, pages 14246--14255, 2023.

\bibitem[Yeshwanth et~al.(2023)Yeshwanth, Liu, Nie{\ss}ner, and Dai]{scannet++}
Chandan Yeshwanth, Yueh-Cheng Liu, Matthias Nie{\ss}ner, and Angela Dai.
\newblock Scannet++: A high-fidelity dataset of 3d indoor scenes.
\newblock In \emph{Proceedings of the IEEE/CVF International Conference on Computer Vision}, pages 12--22, 2023.

\bibitem[Zhang et~al.(2023)Zhang, Rao, and Agrawala]{zhang2023adding}
Lvmin Zhang, Anyi Rao, and Maneesh Agrawala.
\newblock Adding conditional control to text-to-image diffusion models.
\newblock In \emph{Proceedings of the IEEE/CVF International Conference on Computer Vision}, pages 3836--3847, 2023.

\bibitem[Zhao et~al.(2019)Zhao, Meng, Yin, and Sigal]{zhao2019image}
Bo Zhao, Lili Meng, Weidong Yin, and Leonid Sigal.
\newblock Image generation from layout.
\newblock In \emph{Proceedings of the IEEE/CVF Conference on Computer Vision and Pattern Recognition}, pages 8584--8593, 2019.

\end{thebibliography}
}
\maketitlesupplementary

\section{StagePro-V1 dataset}
\begin{figure}[ht]
\centering
\includegraphics[width=8.3cm]{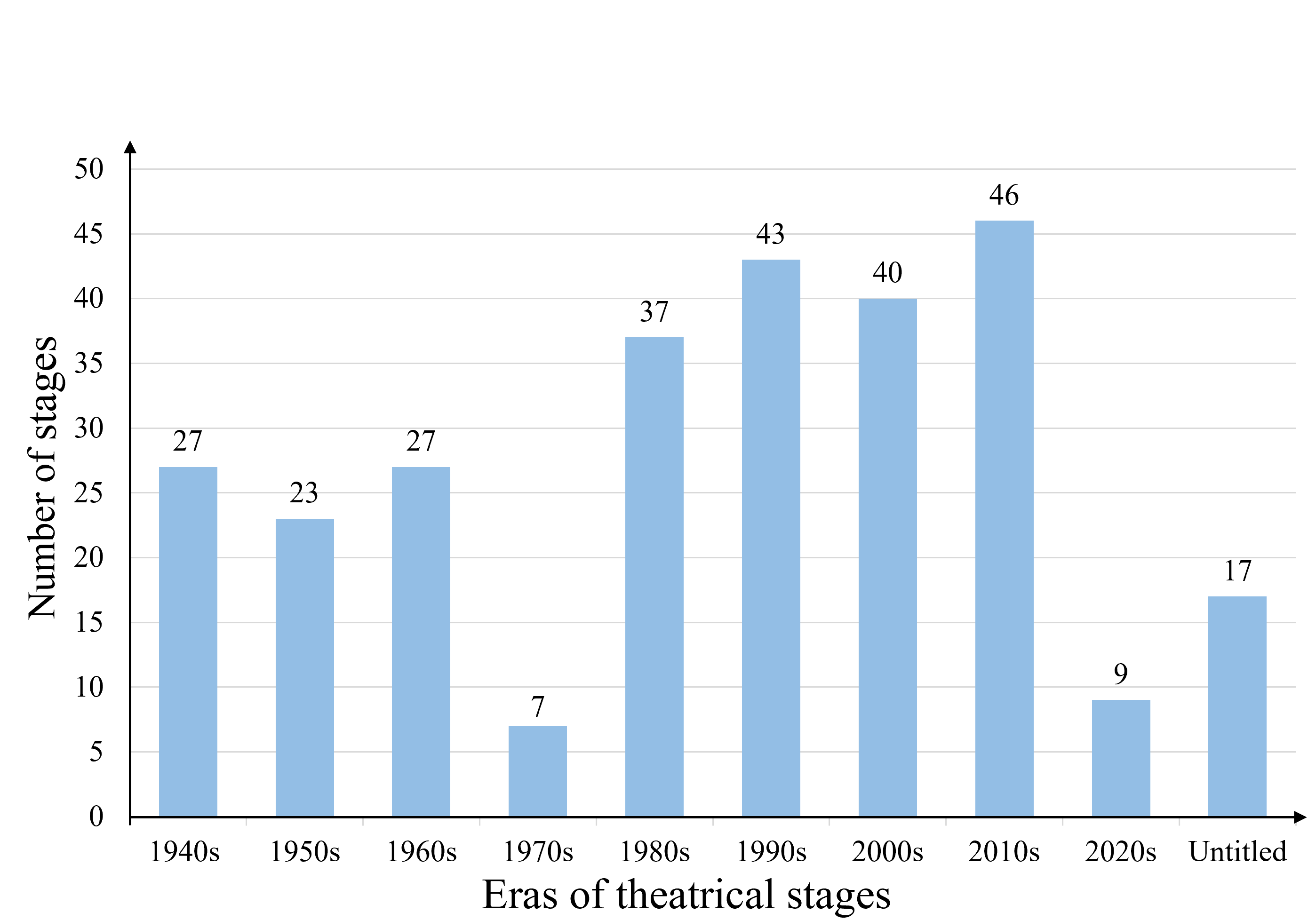}
    \caption{Distribution of stage productions in the dataset across decades, from the 1940s to the 2020s.}
    \label{fig:era}
    \vspace{-0.5cm}
\end{figure}

\begin{figure}[ht]
\centering
\includegraphics[width=8.3cm]{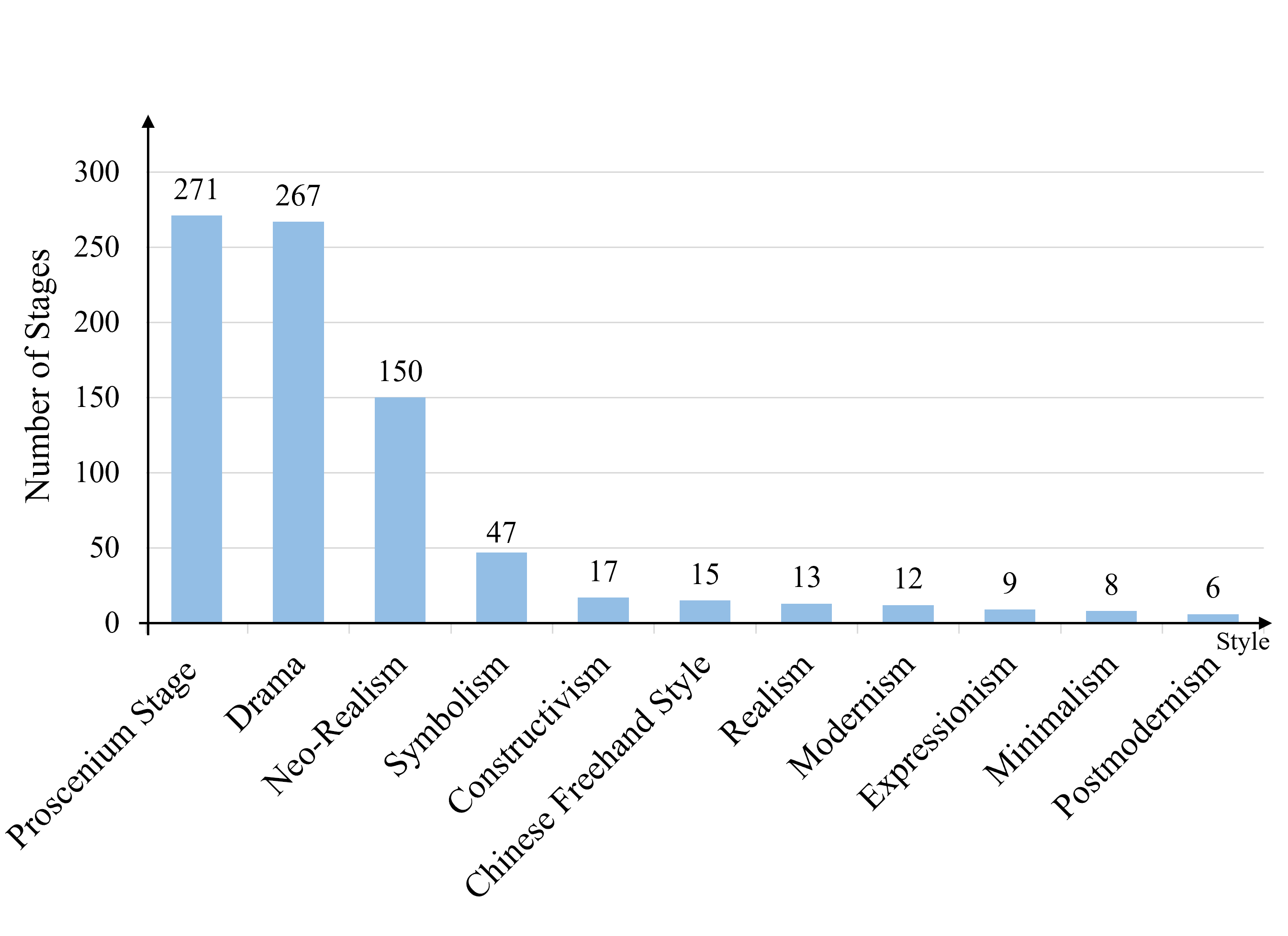}
    \caption{Style distribution in the dataset, where each stage may exhibit multiple styles. The counts represent the frequency of styles across the dataset.}
    \label{fig:style}
    \vspace{-0.5cm}
\end{figure}

\begin{figure}[ht]
\centering
\includegraphics[width=8.3cm]{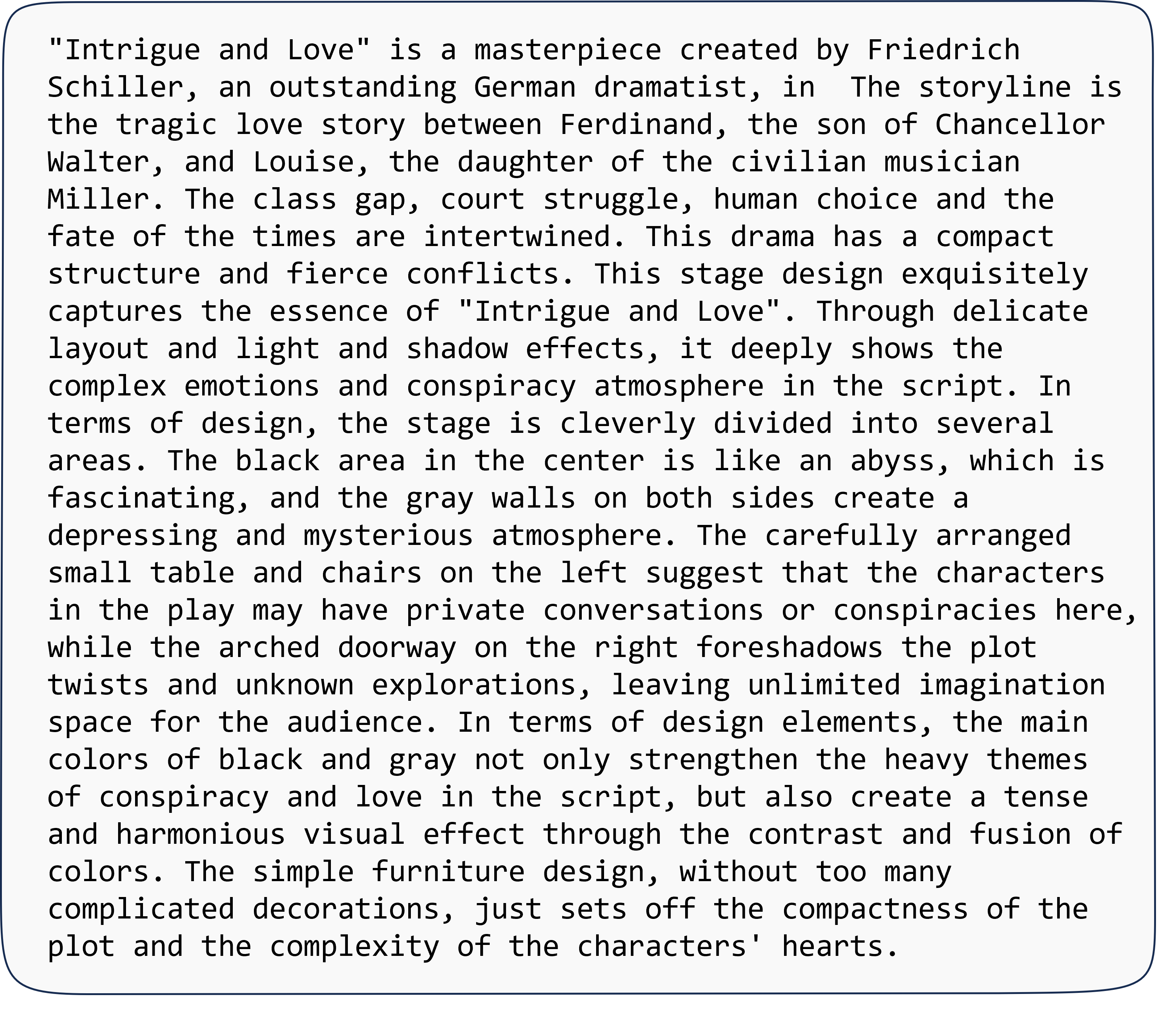}
    \caption{Script example from the StagePro-V1 dataset, showcasing the play \textit{Intrigue and Love} by Friedrich Schiller.}
    \label{fig:script}
    \vspace{-0.5cm}
\end{figure}

Figures~\ref{fig:era} and~\ref{fig:style} provide additional insights into the diversity of the StagePro-V1 dataset. Figure~\ref{fig:era} illustrates the temporal distribution of stage productions in the dataset, spanning from the 1940s to the 2020s, showcasing the historical breadth of the collected scenes. Figure~\ref{fig:style} highlights the style distribution, where stages often incorporate multiple stylistic elements, reflecting the dataset's richness and adaptability for various artistic scenarios. Figure~\ref{fig:script} presents an example of a script from the dataset, featuring the play "Intrigue and Love" by Friedrich Schiller, exemplifying the detailed narrative and descriptive elements available in the dataset. 

\section{Prompts of StageDesigner}

To provide clarity on the detailed implementation of the StageDesigner, we present the specific prompts used in various modules to transform input scripts into stage layouts.

\begin{figure*}[ht]
\hsize=\textwidth
\centering
\includegraphics[width=\textwidth]{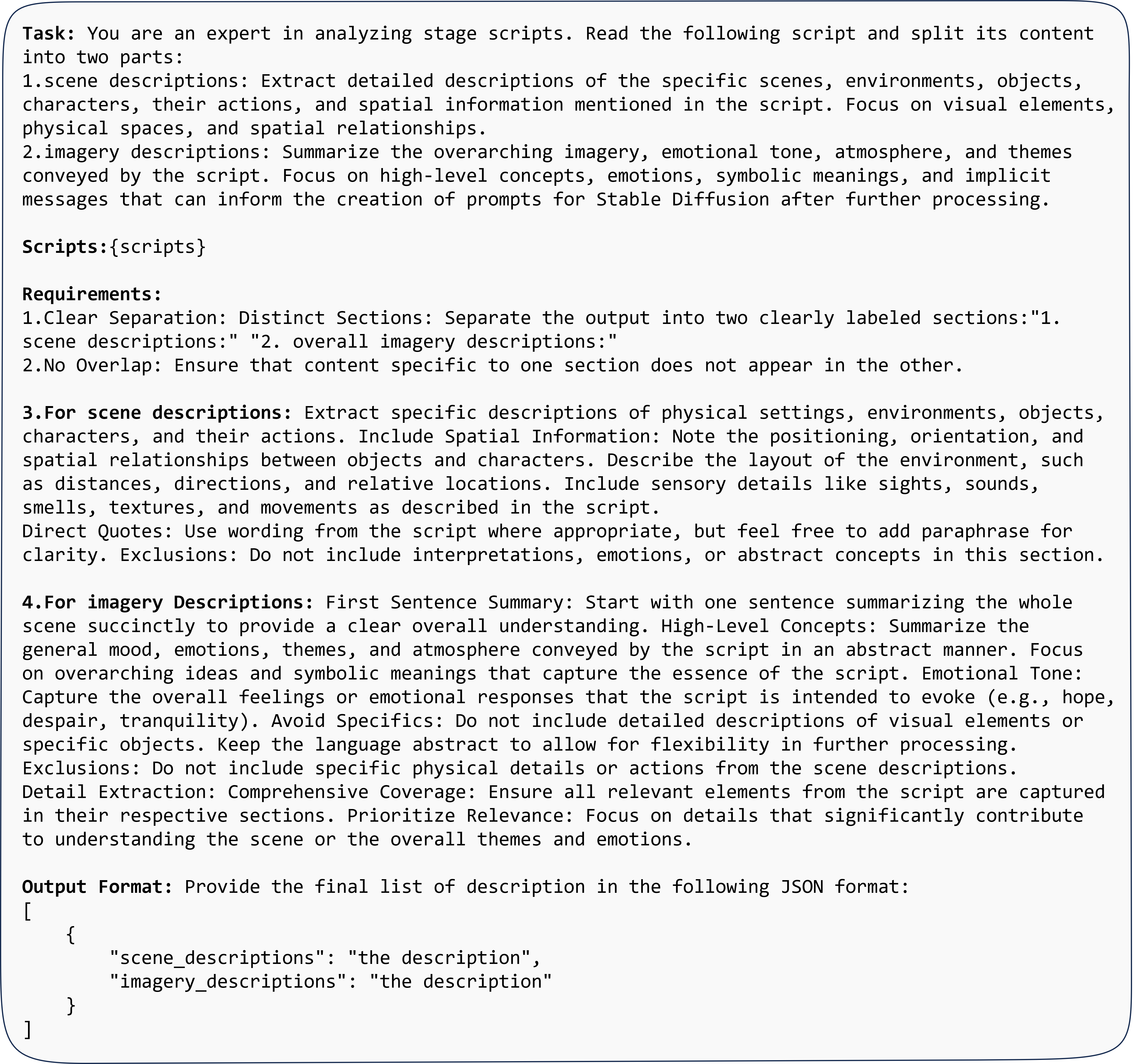}
    \caption{The prompt for Script Analysis Module. Decomposing script into scene description and imagery description.}
    \label{fig: list}
    \vspace{-0.5cm}
\end{figure*}

\begin{figure*}[ht]
\hsize=\textwidth
\centering
\includegraphics[width=\textwidth]{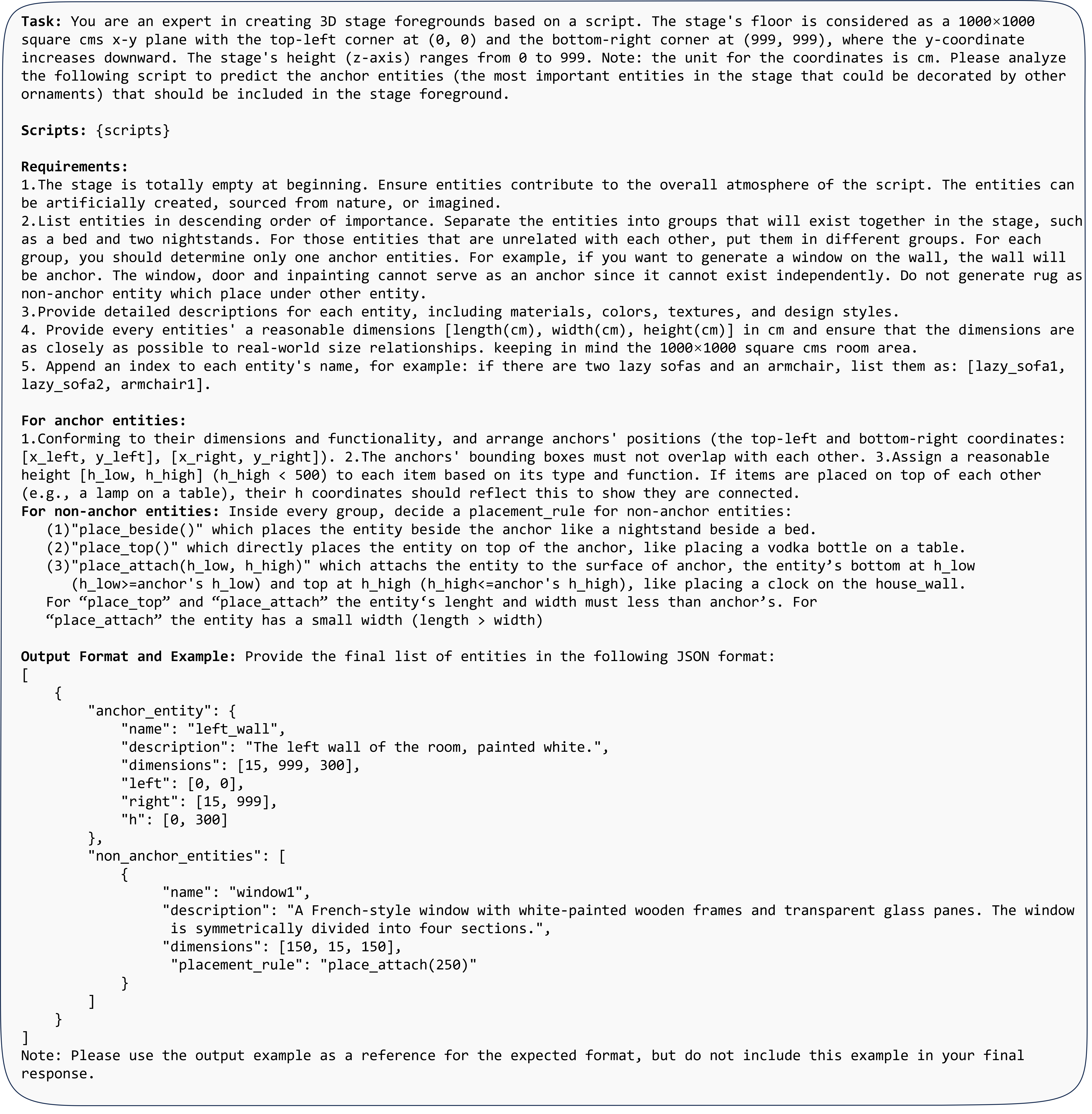}
    \caption{The prompt for anchor and non-anchor entities generation.}
    \label{fig: anchor}
    \vspace{-0.5cm}
\end{figure*}

\begin{figure*}[ht]
\hsize=\textwidth
\centering
\includegraphics[width=\textwidth]{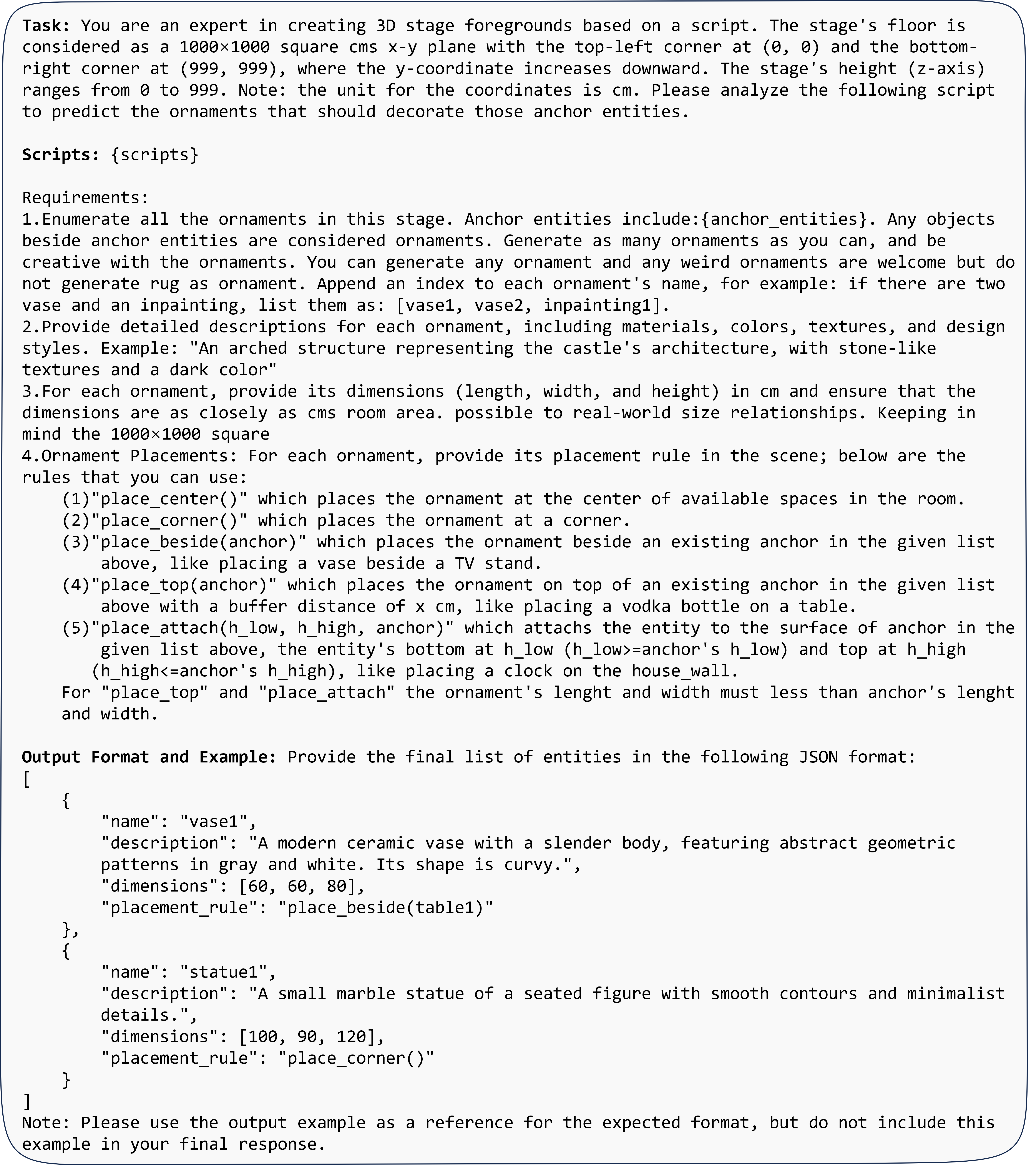}
    \caption{The prompt for ornaments generation.}
    \label{fig: ornaments}
    \vspace{-0.5cm}
\end{figure*}

\begin{figure*}[ht]
\hsize=\textwidth
\centering
\includegraphics[width=\textwidth]{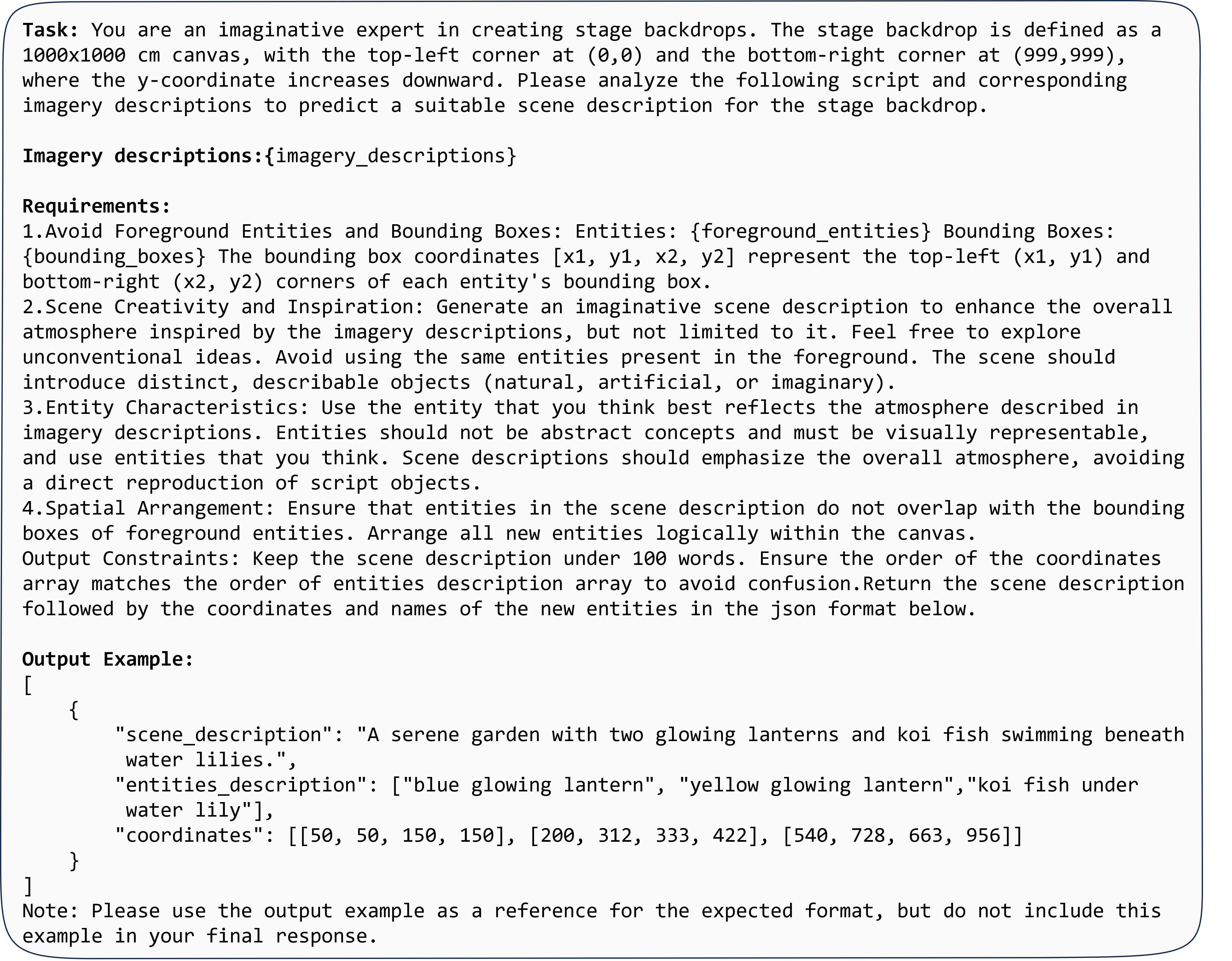}
    \caption{The prompt for layout-controlled background generation.}
    \label{fig: layout}
    \vspace{-0.5cm}
\end{figure*}

\begin{itemize} \item \textbf{Script Analysis Module (Figure~\ref{fig: list})}: The prompt is designed to decompose the input theater script into two distinct parts: scene description, which captures spatial and physical elements, and imagery description, which focuses on the thematic and emotional aspects.

\item \textbf{Anchor and Non-Anchor Entity Generation (Figure~\ref{fig: anchor})}: This prompt guides the generation of primary entities (anchors) and their corresponding dependent entities (non-anchors), ensuring spatial coherence and logical arrangements in the stage layout.

\item \textbf{Ornaments Generation (Figure~\ref{fig: ornaments})}: This prompt is tailored to create ornamental entities that enhance the stage's overall aesthetic, specifying placement rules and ensuring alignment with the scene's design.

\item \textbf{Layout-Controlled Background Generation (Figure~\ref{fig: layout})}: The prompt directs the creation of background elements, emphasizing layout control to avoid occlusion of key foreground elements while maintaining thematic consistency with the script.
\end{itemize}

\noindent These prompts showcase the step-by-step process through which StageDesigner ensures thematic alignment, spatial coherence,and aesthetic richness in the generated stage.

\begin{figure*}[ht]
\hsize=\textwidth
\centering
\includegraphics[width=\textwidth]{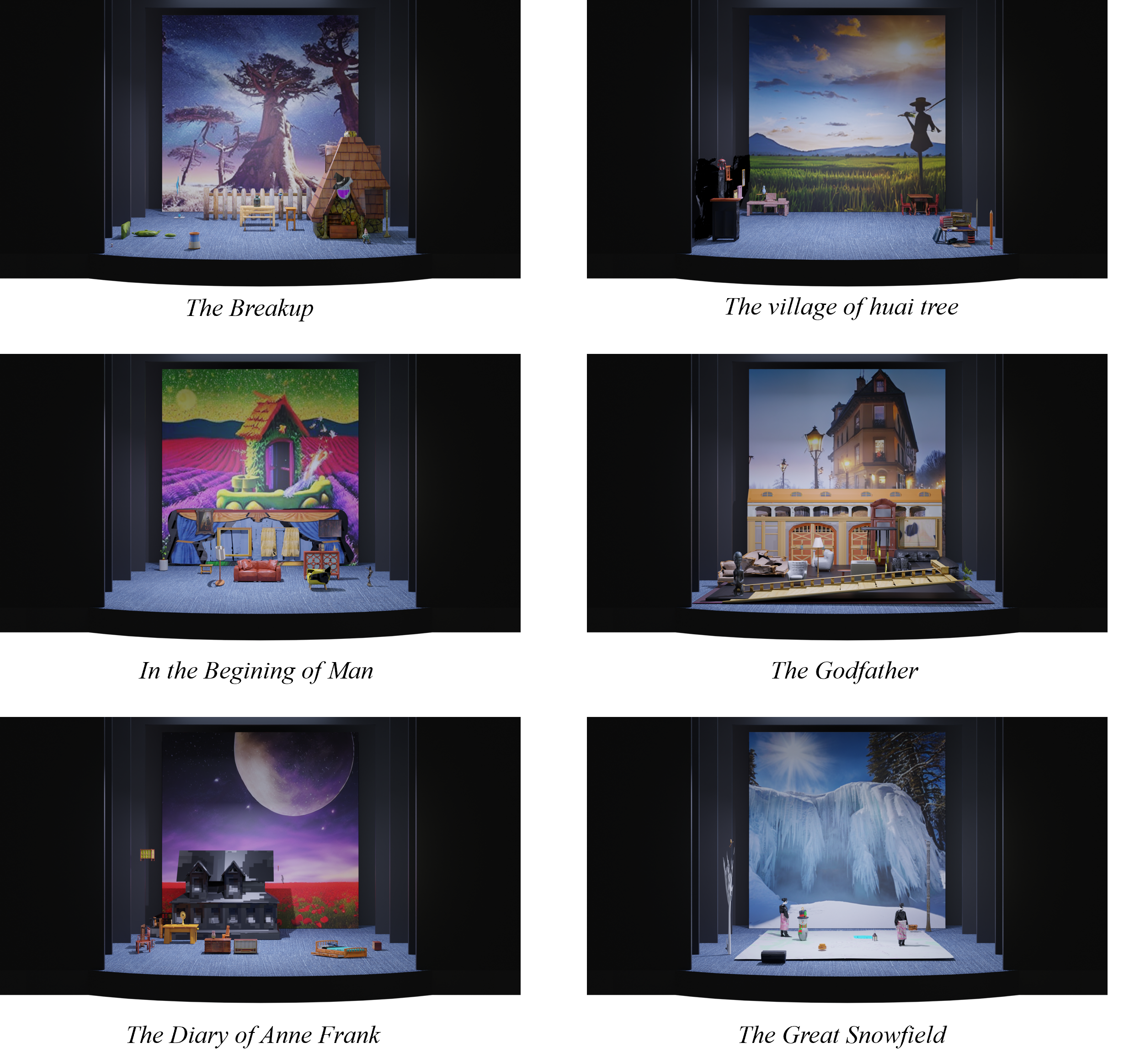}
    \caption{More generated stages by StageDesigner.}
    \label{fig: more1}
    \vspace{-0.5cm}
\end{figure*}

\begin{figure*}[ht]
\hsize=\textwidth
\centering
\includegraphics[width=\textwidth]{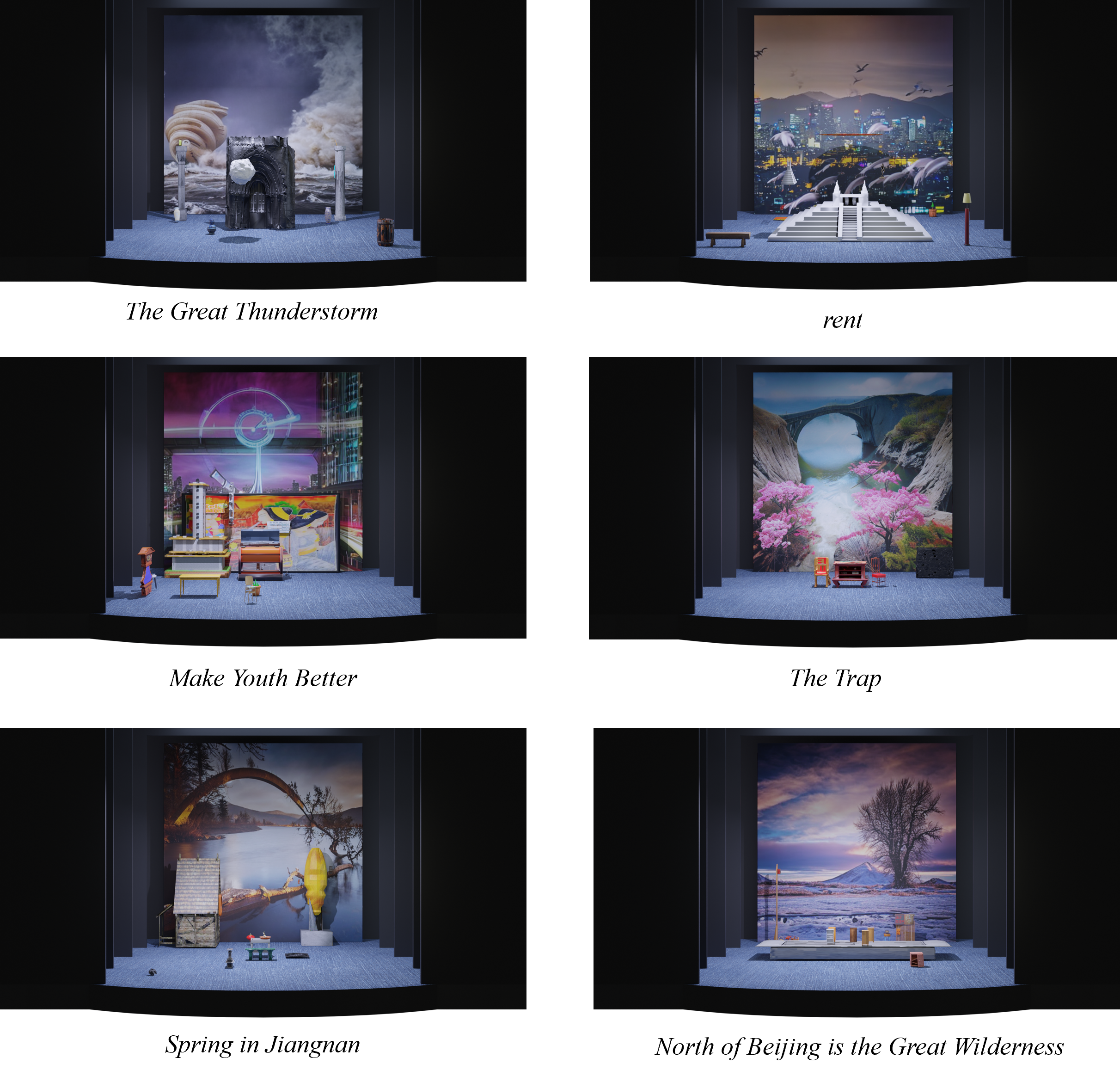}
    \caption{More generated stages by StageDesigner.}
    \label{fig: more2}
    \vspace{-0.5cm}
\end{figure*}


\end{document}